\documentclass[journal]{IEEEtran}
\usepackage{caption}
\usepackage{graphicx}
\usepackage{epstopdf}
\IEEEoverridecommandlockouts
\pdfoutput=1
\usepackage[T1]{fontenc}% optional T1 font encoding
\usepackage{cite}
\usepackage{threeparttable}
\usepackage{amsmath}
\usepackage{times} % assumes new font selection scheme installed
\usepackage{amssymb}  % assumes amsmath package installed
\usepackage{graphicx}
\usepackage{subfigure}
\usepackage{bm}
\usepackage{booktabs}
\usepackage{float}
\usepackage{bigstrut}
\usepackage{multirow}
\usepackage{balance}
\usepackage{url}
\usepackage{bm}
\usepackage{booktabs}
\usepackage{float}
\usepackage{bigstrut}
\usepackage{caption}
\usepackage{amsopn}
\usepackage{epstopdf} % for eps figure using pdflatex
\usepackage[vlined,ruled]{algorithm2e}
\usepackage[table]{xcolor}
\usepackage[mathlines,switch]{lineno}
\usepackage{placeins}
\usepackage{amsbsy}
\usepackage{amsthm}% for the lemma and proof
\usepackage{graphicx}
\usepackage{subfigure}
\usepackage{amsmath}
\usepackage{graphicx}
\usepackage[inkscapelatex=false]{svg}
\usepackage{algpseudocode}
\usepackage{bm}
\usepackage{makecell}
\usepackage{diagbox}
\usepackage{indentfirst}
\usepackage[misc]{ifsym}
\usepackage{amsfonts}
\usepackage{multirow}
\usepackage{array}
\usepackage{booktabs}
\usepackage{url}
\usepackage{CJKutf8}
\usepackage{dblfloatfix}
\usepackage{arydshln}
\usepackage[utf8]{inputenc}
\def\BibTeX{{\rm B\kern-.05em{\sc i\kern-.025em b}\kern-.08em
    T\kern-.1667em\lower.7ex\hbox{E}\kern-.125emX}}

\begin{document}
\title{
	GVD-Exploration: An Efficient Autonomous Robot Exploration Framework Based on Fast Generalized Voronoi Diagram Extraction
}

\author{Dingfeng Chen, Anxing Xiao, Meiyuan Zou, Wenzheng Chi$^{*}$, Jiankun Wang$^{*}$, and Lining Sun
\thanks{* corresponding author}
\thanks{The work was supported by the National Natural Science Foundation of China (K111701022).}
\thanks{Dingfeng Chen, Meiyuan Zou, Wenzheng Chi, and Lining Sun are with  the Robotics and Microsystems Center, School of Mechanical and Electric Engineering, Soochow University, Suzhou 215021, China. ({e-mail: dfchen1105@stu.suda.edu.cn, \{myzou, wzchi, lnsun\}@suda.edu.cn})}
\thanks{Anxing Xiao is with the School of Computing, National University of Singapore, Singapore 117417, Singapore. ({e-mail: anxingx@comp.nus.edu.sg})}
\thanks{Jiankun Wang is with Shenzhen Key Laboratory of Robotics Perception and Intelligence, and the Department of Electronic and Electrical Engineering, Southern University of Science and Technology, Shenzhen, China. (e-mail:  wangjk@sustech.edu.cn)
Jiankun Wang is also with the Jiaxing Research Institute, Southern University of Science and Technology, Jiaxing, China.}
}

% make the title area
\maketitle

\begin{abstract}
Rapidly-exploring Random Trees (RRTs) are a popular technique for autonomous exploration of mobile robots.
However, the random sampling used by RRTs can result in inefficient and inaccurate frontiers extraction, which affects the exploration performance.
To address the issues of slow path planning and high path cost, we propose a framework that uses a generalized Voronoi diagram (GVD) based multi-choice strategy for robot exploration.
Our framework consists of three components: a novel mapping model that uses an end-to-end neural network to construct GVDs of the environments in real time; a GVD-based heuristic scheme that accelerates frontiers extraction and reduces frontiers redundancy; and a multi-choice frontiers assignment scheme that considers different types of frontiers and enables the robot to make rational decisions during the exploration process.
We evaluate our method on simulation and real-world experiments and show that it outperforms RRT-based exploration methods in terms of efficiency and robustness.
\end{abstract}

\def\abstractname{Note to Practitioners}
\begin{abstract}
This paper aims to improve the autonomous exploration performance of the mobile robot and speed up the exploration process.
An efficient and robust autonomous robot exploration framework is proposed.
In contrast to RRT-based autonomous exploration for mobile robot, our approach extracts frontiers based on GVD information gain, independent of random tree growth, effectively mitigating challenges associated with trap space issues.
Diverging from the majority of existing robotic exploration strategies, our robot employs a diverse set of exploration decisions based on various frontiers, rather than relying on a singular decision approach.
In this regard, the autonomous robot prioritizes exploration within local boundaries, while during global exploration, it transforms the exploration task into a TSP problem for optimization, ensuring the robot makes optimal behavioral decisions.
Finally, the feasibility and reliability of the proposed method have been validated through experiments in various simulated and real-world environments. The proposed framework can be applied to autonomous exploration of mobile robots in diverse types of environments.
\end{abstract}

\begin{IEEEkeywords}
Robot Exploration, GVD, Heuristic Frontiers Fusion Extraction, Multi-Strategy Frontier Assignment.
\end{IEEEkeywords}

\section{INTRODUCTION}
\begin{figure}[]
	\centering
	\includegraphics[width=3.3in]{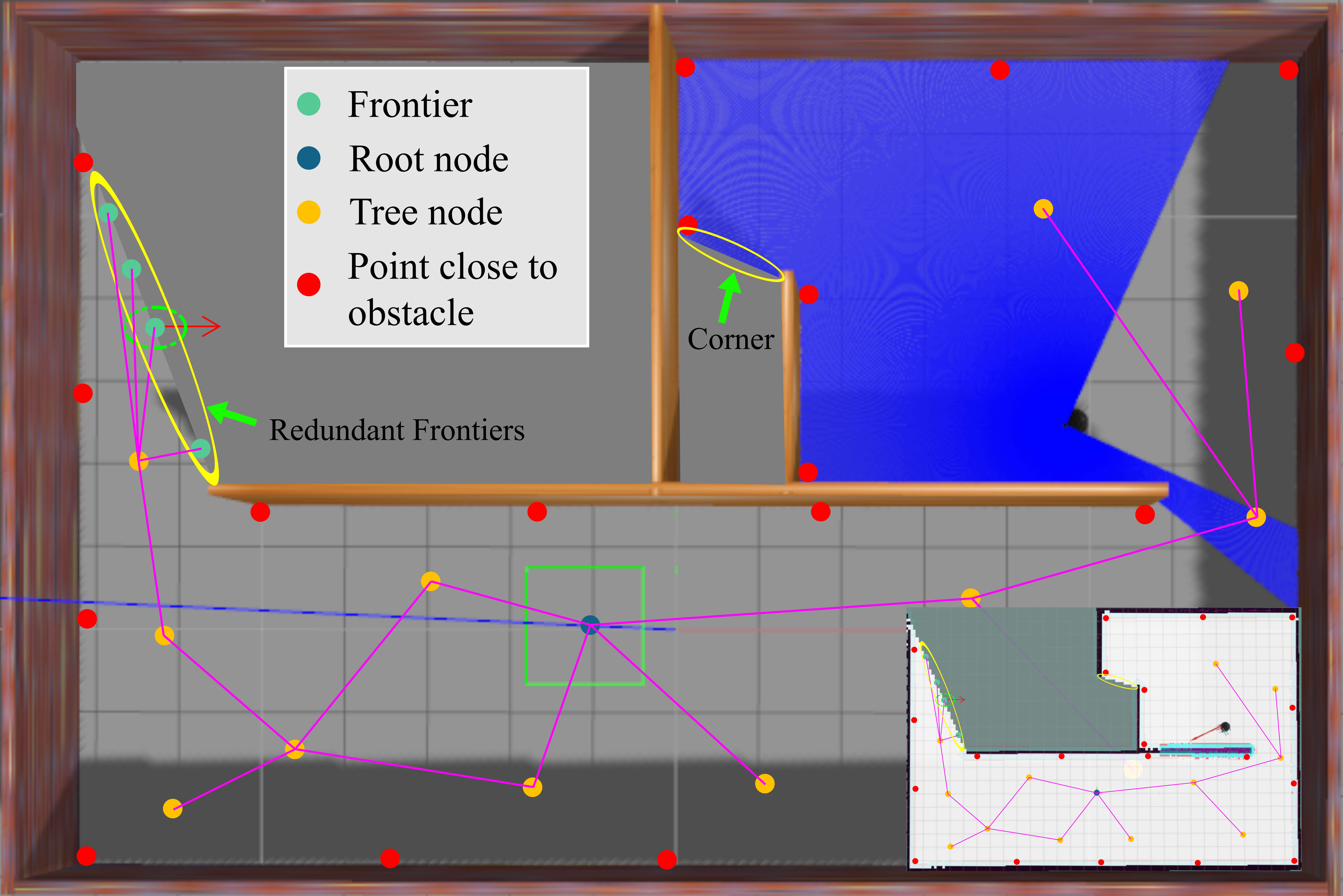}
	\caption{Autonomous robot exploration based on Rapidly-exploring Random Tree. The blue dot is the root of RRT tree and the yellow dots are the RRT nodes reserved by the RRT tree. The green dots are the frontiers extracted by the RRT tree. The frontiers extracted based on the RRT tree are redundant, and the frontiers of those small corners cannot be extracted in time.}
	\label{fig1}
\end{figure}

The goal of autonomous exploration is to enable a mobile robot to navigate an unknown environment without human intervention and build a complete map of it.
The main challenge of this task is to determine the sequence of movement targets that allow the robot to acquire information about the unknown areas in the shortest collision-free path.
A common approach to this problem is the frontier-based exploration method \cite{1,2,3}, which identifies the boundaries between known and unknown regions in the local grid map and guides the robot to explore them. 
However, this method faces two difficulties: how to efficiently and accurately extract frontiers in complex environments, such as mazes, narrow corridors and corners, where the frontiers may be sparse or redundant (see Fig. \ref{fig1}); and how to select frontiers that maximize the information gain and minimize the path cost, while taking into account the global continuity of the frontiers, rather than choosing the locally optimal ones that may lead to inefficient exploration.

We aim to improve the efficiency of autonomous exploration by using the Generalized Voronoi Diagram (GVD) \cite{5} to guide the frontier-based exploration method. The GVD consists of the points that are equidistant to the nearest obstacles \cite{33,34}.
We use the GVD to extract heuristic frontiers and assign them hierarchically to reduce exploration time and path cost.
To speed up the GVD construction, we propose a GVD mapping model that applies pooling and convolution operations on the environment map and the obstacle distance map.
After obtaining the GVD, we merge the heuristic frontiers based on the geometry of the GVD nodes, and evaluate the cost and information gain of each frontier.
For local exploration, we choose the frontier with the lowest cost. For global exploration, we solve the Traveling Salesman Problem (TSP) to find the optimal order of visiting each frontier.
We use the GVD path \cite{32} instead of the Euclidean distance to calculate the path cost, as it considers the connectivity of the environment.
The GVD path can be easily found by connecting the robot pose and frontier to the Voronoi edges and using Dijkstra algorithm.

The contributions of our work can be summarized as follows: 
\begin{itemize} 
	\item A novel GVD mapping model that builds the GVD faster; 
	\item A GVD-based method that extracts heuristic frontiers and reduces redundancy; and 
	\item An effective hierarchical framework for assigning different types of frontiers. 
\end{itemize}

The rest of the paper is structured as follows.
Section \ref{sec:related_work} reviews the related work on robot exploration methods.
Section \ref{sec:method} explains the GVD-based exploration strategy in detail.
Section \ref{sec:results} presents and discusses the experimental results.
Section \ref{sec:conclusions} concludes the paper.

\section{RELATED WORK}
\label{sec:related_work}
Autonomous exploration of mobile robots is an important research direction in the field of robotics.
The most important task of mobile robot autonomous exploration is to determine the next expected movement position of the robot, and finally realize the acquisition of the most unknown and correct environmental information with the shortest collision-free path in the global scope.
Most autonomous exploration schemes are based on the ideas of frontier region (Frontier) and best view (NBV, Next Best View) when determining travel goals.
The frontier-based exploration method \cite{7,8,9,10,11,12} divides the exploration area into known areas and unknown areas through the frontier, and updates the environment map by guiding the robot to move to the frontier.
The NBV exploration method \cite{13} does not consider looking for frontiers, but generates viewpoints in all known spaces, and then compares the comprehensive income to determine the best view at the next moment.
Compared with the frontier-based exploration method, NBV reduces the amount of calculation at the cost of ignoring global information, which may lead to incomplete mapping.
Therefore, NBV method may not be suitable for large multi-channel environments.
In this paper, we propose to adopt the frontier-based exploration method.

In the aspect of frontier extraction, some scholars \cite{14,15,16,17} utilize the image processing technology to extract frontiers from the map.
Yamauchi \cite{16} proposed a frontier-based exploration algorithm based on the image segmentation technology to extract frontiers, and then assigned the nearest frontier to the robot to obtain new environmental information.
In order to reduce the consumption of computing resources, Keida and Kaminka \cite{17} proposed Wavefront Frontier Detector (WFD) and Fast Frontier Detector (FFD) to process the local data instead of the global map data to speed up the frontier extraction.
However, these methods are more suitable for small-scale scenarios.
As the map size increases, the computational complexity will increase exponentially.
To avoid the complex construction of the configuration space, Umari \textit{et al.} \cite{18} employed Rapidly-exploring Random Tree (RRT) to extract frontiers and the nodes where the RRT tree grows to the boundary are marked as prospective frontiers.
However, due to the randomness of the RRT algorithm, the frontiers at the map corner can not be extracted in time, resulting in backtracking in the exploration process.
Moreover, the robot may trapped in some complex environments, resulting in the failure of robot autonomous exploration.

In the field of frontier selection, a popular approach has been designed according to the information theory \cite{19,20,21,22,23}, aiming to reduce the entropy of the map at the fastest speed and maximize the information gain.
However, majority of these existing methods rely on greedy strategies, where efficiency is limited due to the methods are myopic.
Besides, a common approach is based on utility function.
Yamauchi \textit{et al.} \cite{16} assigned the closest frontier to the robot as the exploration target.
Moorehead \textit{et al.} \cite{24} proposed using multiple sources of information for two planning strategies: one is random walk using sensor information and the other is greedy search.
Mei \textit{et al.} \cite{25} combined the direction information of the robot in the utility function, taking into account the energy and time loss of the robot going to the frontier.
Stachniss \textit{et al.} \cite{26} presented a decision-theoretic framework to evaluate the cost by considering the uncertainty and expected information gain.
Umari \textit{et al.} \cite{18} clustered frontiers and improved information gain in the utility function, taking into account the unknown number of cells within a predetermined radius.
Liu \textit{et al.} \cite{12} combined the semantic information of the environment to prioritize the heuristic frontiers.
Dieter Fox \textit{et al.} \cite{27} proposed computing the cost and utility of each robot target frontier, and then use a linear program solver to find the optimal assignment.
However, most of the frontiers selected by this method are not globally optimal in that the impact of the exploration order between each boundary point on the exploration efficiency is not fully considered.
In order to solve this problem, some scholars \cite{28,29,30} proposed employing Traveling Salesman Problem (TSP) to get the sequence of visiting all frontiers.

In addition, when evaluating the path cost of the robot current pose to the target frontier, two methods are usually adopted: one is the Euclidean distance, and the other is the specific path length through the path planning method.
Umari \textit{et al.} \cite{18} utilized Euclidean distance to calculate the path cost.
However, the geometric characteristics of the environment are ignored, which can easily lead to inaccurate evaluation of the optimal frontier, resulting in the phenomenon of exploration backtracking.
Dieter \textit{et al.} \cite{27} computed the minimal path cost efficiently by A$*$ search.
However, the increasing of the environment will lead to an increase in the number of frontiers, which increases the computational complexity and causes the robot to be trapped in place due to time-consuming planning.
%As a result, the robot stays in the same area for too long and the exploration trajectory has a large redundancy, which leads to the irregularity of robot exploration.
In this work, we compute the cost path %efficiently by Dijkstra \cite{31} searching 
on the GVD instead of the whole map, which reduces the computational consumption.
GVD can represent the geometric space of the environment, and the generated path is safer and more suitable for the actual autonomous exploration environment.

%
%In this work, the GVD is constructd and employed to extract the frontiers.
%We design a hierarchical exploration framework to cope with different types of frontiers.
%For local frontiers, we propose utilizing a utility function for optimal frontier assignment.
%For global frontiers, we advocate using the TSP solution strategy to get the sequence of %visiting all frontiers.
%In addition, we utilize GVD path for path cost calculation.
%We conduct a variety of experiments from simulated and real-world scenarios to verify the effectiveness of our work.

\section{METHODOLOGY}
\label{sec:method}
In this paper, we propose a GVD-based exploration method to improve the efficiency of the robot exploration.
%As shown in Fig. \ref{fig:},
The GVD-based exploration method includes three modules: a novel GVD construction mapping module, a GVD-based heuristic frontiers extraction module and a multi-strategy frontier assignment module.
When the robot starts to explore, it first builds the local environment map around it and constructs the GVD of the current map.
Then, by judging unknown grids within the GVD radius, heuristic frontiers of the environment are extracted and fused.
The multi-strategy assignment module assigns different types of heuristic frontiers by utilizing GVD path instead of Euclidean path or planning path.
After the robot decides the optimal frontier, the robot utilize the ${move\_base}$ framework for motion planning.
This process repeats until the robot explores the entire environment.

\begin{figure}[]
	\centering
	\includegraphics[width=3.3in]{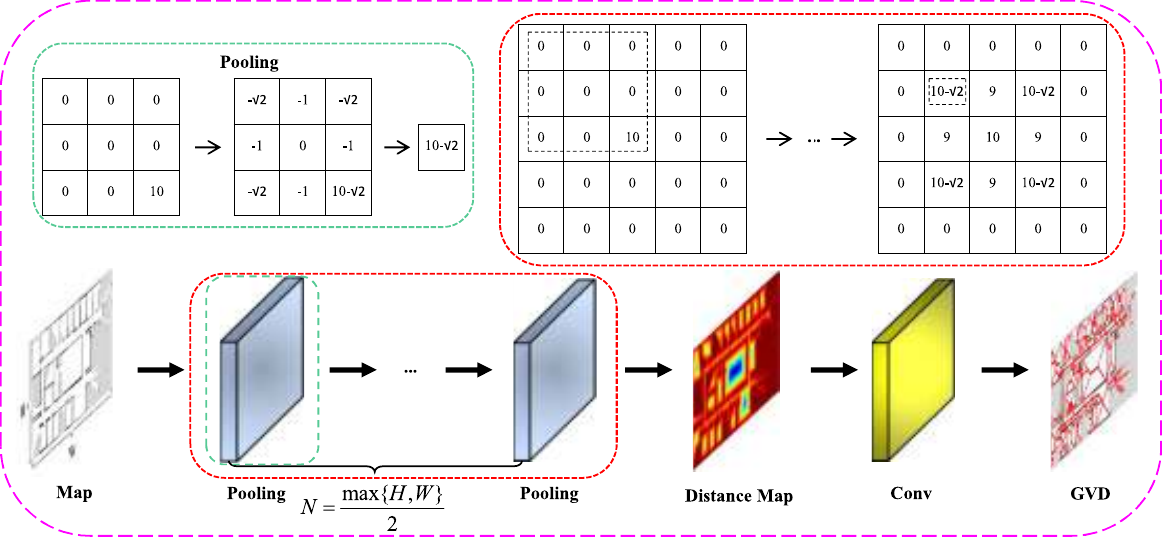}
	\caption{The GVD construction mapping module.}
	\label{fig2}
\end{figure}

\subsection{A Novel GVD Construction Mapping Module}
\label{sec:GVD Generation}
%The GVD can concisely represent the salient geometry of the environmental map
The GVD maximizes the gap between obstacles, and in addition, it can concisely represent the geometry of the environment map.
Traditional GVD extraction needs to calculate the distance from points in free space to obstacles, resulting in high computational complexity.
In the process of autonomous exploration of the robot, it is necessary to extract the environmental features in real time to speed up the exploration of the robot.
In order to speed up the Generalized Voronoi Diagram of the environment map, we use a pre-built mapping model to replace the previous mathematical model \cite{6}.
Our mapping model consists of $N$ pooling layers and a convolution layer, as shown in Fig. \ref{fig2}.
The pooling layer performs pooling operations on the environment map to obtain an obstacle distance map.
The convolution layer performs a convolution operation on the obstacle distance map to obtain the Generalized Voronoi Diagram of the environment map.
Below we elaborate on our model.

%We preprocess the environment map obtained by the robot using the sensor model, converting the image into a binary image.
In the preprocessing stage, the environment map is converted into a binary image, where the pixel value of 1 indicates the obstacle or the unknown area, and a pixel value of 0 indicates the free area.
For the distance from the pixel point in the free area to the obstacle, we use the method of expanding the boundary line of the obstacle outward.
The closed curve obtained by expanding the boundary line of the obstacle by one pixel step each time is recorded as the target baseline, and the distance from each pixel point on the target baseline to the obstacle can be determined by the number of expanded pixel steps.
For the pooling operation, we first need to determine the number of pooling layers.
Multiple pooling operations can improve the accuracy of the Voronoi map of the environment map.
We determine the number of pooling layers according to the size of the environment map.
The formula for determining the number of pooling layers $N$ is given as follows:
\begin{equation}
	\label{eq:revenue_function1}
	\begin{aligned}
		N = \frac{L_{max}}{2}. \\
	\end{aligned}
\end{equation}
Here, ${L_{max}}$ is the maximum of the length and width of the environment map.
When pooling the environment map, first we set the pooling window to 3x3 and the moving step to 1.
Then, the distance offset operation is performed on the image area at the pooling window through a preset distance offset operator.
The image area $X$ after distance offset transformation is:
\begin{equation}
	\label{eq:revenue_function2}
	\begin{aligned}
		X = Ax + B, \\
	\end{aligned}
\end{equation}
where $X$ is the image area after distance offset transformation, $A$ is the identity matrix, and $B$ is the offset matrix.
Here, the offset matrix $B$ is:
\begin{equation}
	\label{eq:revenue_function3}
	\begin{aligned}
		B = \left [\begin{array}{ccc}
			-\sqrt{2} &-1   & -\sqrt{2} \\
			-1 & 0  & -1 \\
			-\sqrt{2} & -1 &-\sqrt{2} \\
		\end{array}\right]
	\end{aligned}
\end{equation}
After the distance offset transformation is performed on the image area, the value corresponding to each pixel in the image area can indicate the distance between the pixel and the obstacle.
The maximum value in the image area transformed by the distance offset is determined as the maximum pooling value of the environment map at the pooling window.
After the pooling window traverses all the pixels in the environment map, the corresponding obstacle distance map can be obtained.

After the pooling operation, the convolution operation is performed on the obstacle distance map to extract image edge features.
Since the distance from the pixel point to the obstacle in the obstacle distance map is gradient, there is a distance extreme edge locally in the obstacle distance map, therefore we use the Laplacian operator as a convolution kernel to perform a convolution operation on the obstacle distance map.
The Laplacian operator $L$ is defined as:
\begin{equation}
	\label{eq:revenue_function4}
	\begin{aligned}
		L = \left [\begin{array}{ccc}
			0 &1   & 0 \\
			1 & -4  & 1 \\
			0 & 1 &0 \\
		\end{array}\right].
	\end{aligned}
\end{equation}
After obtaining the local distance extreme edge of the obstacle distance map, the local distance extremum edge is used as the edge of the Voronoi diagram, and multiple local distance extremum edges in the obstacle distance map are connected to obtain the Voronoi diagram of the environment map.

%\begin{figure}[]
%\centering
%\includegraphics[width=3.3in]{pic/M21unfused.eps}
%\caption{Extracted redundant heuristic frontiers.}
%\label{fig3}
%\end{figure}

%\begin{figure}[]
%\centering
%\includegraphics[width=3.3in]{pic/M22fused.eps}
%\caption{Local and global heuristic frontiers extraction.}
%\label{fig4}
%\end{figure}

\begin{figure}[]
	\centering
	\subfigure[Redundant heuristic frontiers extraction.]
	{
		\centering
		\includegraphics[width=3.3in]{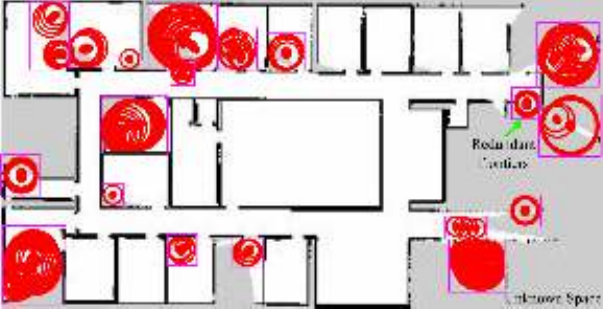}
	}
	\subfigure[Local and global heuristic frontiers fusion extraction.]
	{
		\centering
		\includegraphics[width=3.3in]{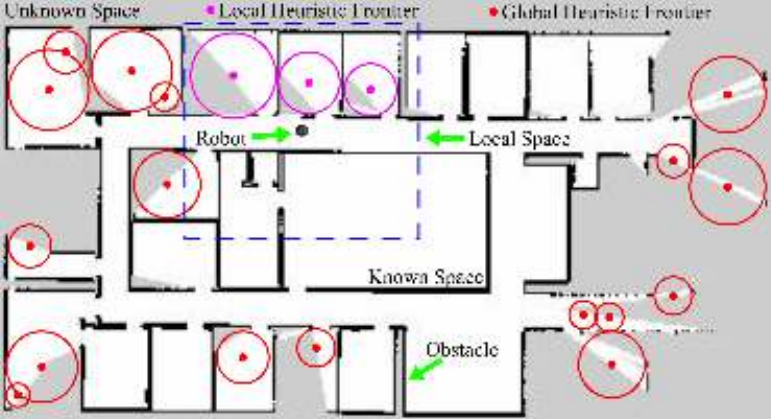}
	}
	\caption{GVD-based Heuristic Frontiers Extraction. (a) The red dots are extracted heuristic frontiers. The red circle is the corresponding node radius. The extraction of heuristic frontiers can result in redundancy due to the existence of multiple GVD nodes with unknown grids within the radius of the same boundary. (b) The purple dots are extracted local heuristic frontiers near the robot and the red dots are extracted global heuristic frontiers. The circle is the corresponding node radius. Fusion extraction of heuristic frontiers can reduce redundancy at the same boundary.}
	\label{fig3}
\end{figure}

\subsection{GVD-based Heuristic Frontiers Fusion Extraction}
\label{sec:Heuristic Frontiers Extraction Module based on GVD}
The GVD has the geometric information of the environment map.
In this work, we utilize the GVD to extract frontiers instead of the whole environment map to reduce the consumption of computing resources.
We define the nodes extracted from GVD as the heuristic frontiers.
Here, we define the GVD node set as $G$ and denote a node on GVD as $G_i$ and the corresponding radius as $r_i$.
In our work, the environment map is represented as an occupancy grid.
With different grid values, the environment map is divided into three states, 0 for free, 100 for occupied and -1 for unknown.
As shown in Fig. \ref{fig3}, heuristic frontiers can be easily extracted by judging whether there is an unknown grid within the radius of the GVD node.
The structural properties of GVD nodes ensure that the node radius will not cross obstacles when determining unknown grids.
Relying on the individual radius of nodes rather than a fixed value ensures the accuracy of heuristic frontiers extraction.
%The GVD node radius guarantees that it will not cross obstacles to judge unknown grids, which ensures the accuracy of heuristic frontiers extraction.

Here, we propose the use of two versions of heuristic frontiers extraction methods based on GVD:
i.) a local heuristic frontiers extraction method, and ii.) a global heuristic frontiers extraction method.
However, we can find that there are many redundant GVD nodes near the same boundary, as shown in Fig. \ref{fig3} (a).
Redundant GVD nodes lead to a time-consuming extraction process and increase the computational complexity of exploration planning.
Hence, we propose a heuristic frontiers fusion extraction algorithm to reduce redundant representations.
%The details related to these are provided below.

%Here, we propose a heuristic frontiers fusion extraction algorithm to reduce redundant representations.
%We advocate the use of two versions of heuristic frontiers extraction method based on GVD:
%i.) a local heuristic frontiers extraction method, and ii.) a global heuristic frontiers extraction method.
%The details related to these are provided below.

\subsubsection{Local Heuristic Frontiers Fusion Extraction Method}
%\label{sec:Local Heuristic Frontier Points Extraction Method}
The local heuristic frontiers fusion extraction algorithm is shown in Algorithm \ref{algorithm1}.
First, we subscribe to the pose of the robot in the environment in real time and build the local environment map of $M$ $*$ $M$ size centered on the robot pose.
The local heuristic frontiers fusion extraction method only processes GVD nodes located in the local environment map.
We define the local GVD node set as $G_{local}$.
This method ensures that the heuristic frontiers closest to the robot itself are extracted, while reducing the consumption of computing resources.
The GVD node with a larger radius usually has richer environmental information, so the local heuristic frontiers fusion extraction starts from the GVD node with the largest radius.
If the number of unknown grids within the radius of the GVD node $G_{max_i}$ is greater than the predefined threshold, it is considered that there is a heuristic frontier worthy of robot exploration near the current GVD node $G_{max_i}$.
To reduce redundant representations, we fuse all GVD nodes within the current GVD node radius $G_{max_i}$.
Otherwise, the current GVD node $G_{max_i}$ is deleted from the local GVD node set $G_{local}$.
This process repeats until the local GVD node set $G_{local}$ is empty.
Local heuristic frontiers are displayed in Fig. \ref{fig3} (b).
The fusion extraction operation reduces the redundant representation of the same boundary, which speeds up the extraction process and reduces the computational complexity of later optimal frontier selection for robot exploration.

\begin{algorithm}[t]
	\LinesNumbered
	\caption{The Local Heuristic Frontiers Fusion Extraction}
	\label{algorithm1}
	\KwIn{The map $M$, local GVD node set $G_{local}$, threshold for the number of unknown grids $\delta$.}
	\KwOut{The local heuristic frontier points set $F_{local}$.}
	%Create an $M$ $*$ $M$ rectangular frame around the center of robot pose $x_{robot}$;\\
	%$G_{local}$ is a local set that all GVD nodes $G$ locating in the rectangular frame;\\
	\While {$G_{local}$ $is$ $not$ $empty$}
	{
		$max_r$ $=$ {$arg$ $max$ ($r_i$)}, $\forall$ $G_i$ $\in$ $G_{local}$; \\
		$num$ is the number of unknown grids within the radius of GVD node $G_{max_r}$;\\
		\If {$num$ $>$ $\delta$}
		{
			add $G_{max_r}$ to $F_{local}$; \\
			\For {all $G_i$ within the $G_{max_r}$ radius $r_{max_i}$}
			{
				remove $G_i$ from $G_{local}$; \\
			}
		}
		\Else
		{
			remove $G_{max_r}$ from $G_{local}$;
		}
	}
	\Return  $F_{local}$;
\end{algorithm}

\subsubsection{Global Heuristic Frontiers Extraction Method}
The global heuristic frontiers fusion extraction algorithm is identical to the local heuristic frontiers fusion extraction algorithm except that the input and output are different.
The input to the global heuristic frontiers fusion extraction algorithm is the whole GVD node set $G$ and the output is the global heuristic frontiers set $F_{global}$.
The global heuristic frontiers fusion extraction algorithm aims to extract all potential frontiers in the environment map, thus guaranteeing the execution of robot autonomous exploration.

In current work, robot autonomous exploration is limited to a single mobile robot, hence, the heuristic frontiers extraction module needs to perform both the local and global heuristic frontiers extraction method.
In future work, we can extend single-robot autonomous exploration to multi-robot collaborative exploration.
Each robot can independently extract the local heuristic frontiers and have access to the global heuristic frontiers of a master robot, which will help reduce computational complexity.

\subsection{Multi-strategy Frontier Assignment Module}
\label{sec:GVD-based Optimal Frontier Assignment Framework}
\begin{figure}[]
	\centering
	\subfigure[]
	{
		\centering
		\includegraphics[width=0.22\textwidth]{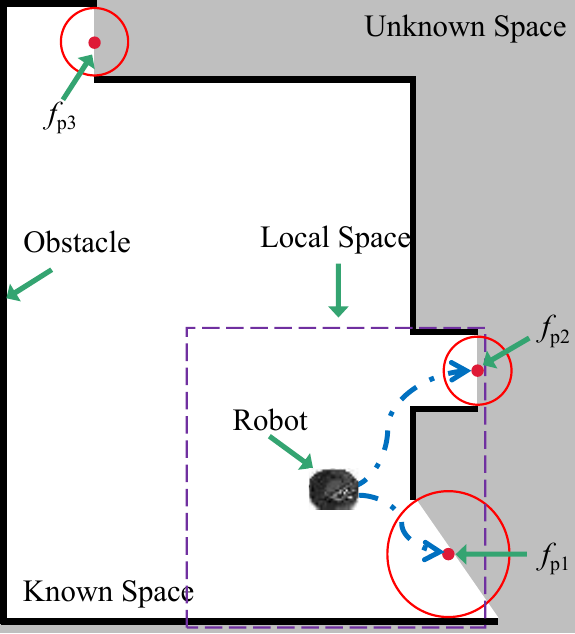}
	}
	\subfigure[]
	{
		\centering
		\includegraphics[width=0.22\textwidth]{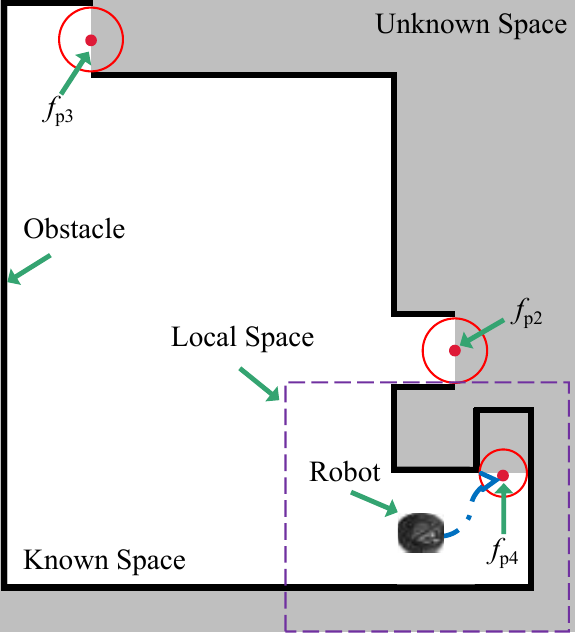}
	}
	\caption{Illustration of the decisions at different timestamps. (a) $f_{p1}$ and $f_{p2}$ are the real-time extracted local heuristic frontiers and $f_{p3}$ is the extracted global frontier. At this time, the highest priority of the frontier is $f_{p1}$ or $f_{p2}$, and $f_{p3}$ is the lowest priority. (b) $f_{p4}$ is the real-time extracted local heuristic frontier, and $f_{p2}$ is the extracted local heuristic frontier. At this time, the priority of the frontier $f_{p4}$ is highest and $f_{p3}$ is lowest. $f_{p2}$ is the priority second best.}
	\label{fig5}
\end{figure}

When the heuristic frontiers of the environment are received, it is necessary to select an optimal frontier in combination with the environmental information for task assignment of the robot, so as to speed up the construction of the environment map.
In this work, we design a multi-strategy frontier assignment framework for multi-type frontiers.
Based on the timestamp information of the robot, we categorize local frontiers into two types: real-time local frontiers and reserved local frontiers.
The multi-strategy frontier assignment framework has the following three strategies, i.) real-time local frontier assignment strategy, ii.) local frontier assignment strategy, and iii.) global frontier assignment strategy.
%The execution of the three strategies has a sequence, in Fig. \ref{fig5}.
As shown in Fig. \ref{fig5}, these three strategies are executed in sequence.
The real-time local frontier assignment strategy has the highest priority.
In addition, we prefer to explore all the frontiers near the robot first, so local frontier assignment strategy is less optimal.
When an area is fully explored, the robot implement the global frontier assignment strategy to ensure the construction of the global map.
For better illustration, we define the terms of frontiers set used in the process as follows:

%\newtheorem{myDef}{Definition}
%\begin{myDef}
%\label{myDef1}
$V_{current}$: a node set that stores the real-time extraction of local frontiers by the local heuristic frontier extraction method.
%\end{myDef}

%\begin{myDef}
%\label{myDef2}
$V_{local}$: a node set that stores all local frontiers extracted by the local heuristic frontier extraction method.
%\end{myDef}

%\begin{myDef}
%\label{myDef3}
$V_{global}$: a node set that stores all global frontiers extracted by the global heuristic frontier extraction method.
%\end{myDef}

\subsubsection{Real-time Local Frontier Assignment Strategy}
%\label{sec:Real-time Local Frontier Point Assignment Framework}
When the real-time local frontiers are received, the robot adopts the real-time local frontier assignment strategy.
We design a cost function to each frontier ${C}$ with the following formula:
\begin{equation}
	\label{eq:revenue_function}
	\begin{aligned}
		C{(x_{r},x_{f})}= G{(x_{r},x_{f})}- t{(x_{r},x_{f})} \ast I{(x_{f})},\\
	\end{aligned}
\end{equation}

\begin{equation}
	\label{eq:canshu_function}
	\begin{aligned}
		t{(x_{r},x_{f})} =
		\begin{cases}  0,& \text{$G{(x_{r},x_{f})}$ < ${\lambda}$}\\
			{\gamma},& \text{${\gamma}$ > 1} \end{cases}
	\end{aligned}
\end{equation}
where $x_{r}$ is the robot current pose, and $x_{f}$ is the extracted frontier.
$G{(x_{r},x_{f})}$ is the distance from the robot current pose to the frontier, and $t{(x_{r},x_{f})}$ is the condition parameter function.
$I{(x_{f})}$ is the number of unknown grids within a fixed radius centered on the frontier $x_{f}$.
Within Eq. (\ref{eq:canshu_function}), ${\lambda}$ is a user-defined parameter used to determine whether to consider the information gain of the frontier.
${\gamma}$ is a constant to increase the influence of the information gain of the frontier itself on the cost function ${C}$.
For each frontier, the cost is calculated by using Eq. (\ref{eq:revenue_function}).
The frontier with the lowest cost is assigned to the robot for exploration.
In this strategy, we prioritize selecting the frontier closest to the robot current pose to ensure that the robot can fully explore one direction in order.
Therefore, we can set the parameter ${\lambda}$ large to ignore the influence of information gain.
As mentioned above, we utilize GVD path instead of Euclidean distance or planning path for path cost calculation.
The local frontiers are extracted in real time during the robot exploration process, once the local frontiers extracted, the robot will immediately choose this strategy.
The strategy ends when $V_{current}$ is empty.
This strategy can ensure that the robot can explore completely along a branch, so as to reduce the backtracking phenomenon caused by robot incomplete exploration.

\subsubsection{Local Frontier Assignment Strategy}
When $V_{current}$ is empty, the robot adopts the local frontier assignment strategy.
This strategy preserves all unexplored local frontiers during the robot exploration.
Before the strategy is executed, the information gain calculation is performed on all the preserved frontiers, and those frontiers without information gain are deleted.
This strategy ensures that after a path branch is fully explored, the robot can turn to other branches recorded by the robot for exploration.
This strategy has the same cost function as the real-time local frontier assignment strategy.
The frontier with the lowest cost is assigned to the robot for exploration.
Frontier with large information gain can speed up the construction of the environment map and reduce the exploration decision of the robot.
Therefore, the robot needs to comprehensively consider the path cost and information gain to speed up the exploration of the area.
In Eq. (\ref{eq:canshu_function}), we set set the parameter ${\lambda}$ to 0, and appropriately increase the parameter ${\gamma}$ to avoid trajectory overlapping.
%This strategy is designed to allow the robot to choose the remaining path branches to explore after a path branch is fully explored.
When the robot explores towards a new branch, if real-time local frontier extracted, the robot turns to the real-time local frontier assignment strategy; otherwise, the robot continues to adopt this strategy until $V_{local}$ are empty, which also means that all branches in the area have been explored completely.

\subsubsection{Global Frontier Assignment Strategy}
When $V_{current}$ and $V_{local}$ are empty, it means that the current sub-area has been fully explored, and the robot needs to turn to other unexplored sub-areas.
At this time, the robot adopts the global frontier assignment strategy.
This strategy records all unexplored global frontiers during the robot exploration.
Similarly, the robot preprocess these frontiers first, and delete the frontiers without information gain.
For the robot global exploration, it is not enough to consider only the lowest cost at present, but it is also necessary to determine an optimum sequence of these frontiers that minimizes the exploration time and path length.
Thus, we can transform the traversal problem of target frontier into the solution of the TSP problem.
TSP problem is a no polynomial-time known solution problem (NP hard problem).
Ant Colony Optimization (ACO) can converge to the global optimum, so this paper adopts ant colony algorithm to find the optimal sequence of frontiers.
The main steps of the ACO algorithm can be summarized as follows:
\begin{itemize}
	\item [1)]
	Set parameters and initialize the pheromone trails.
	\item [2)]
	Ants search for the path and update pheromones.
	\item [3)]
	Iterate this process until the iteration termination condition is met.
\end{itemize}

In this work, the TSP problem is designed to measure the path cost between frontiers and a cost function is given to reward each frontier.
The TSP problem is originally designed to guide the robot return to the initial start point at the end, however, in the exploration problem, this is not necessary.
In order to eliminate the impact of this problem, we set the robot pose as a virtual start, and set the path cost to 0 for all frontiers to the virtual start.
We assume that the total number of frontiers and virtual start is $N$, and the number of possible  pairs becomes ${N}$${(N-1)}$/2.
When the environment is large, it is unreasonable to use path planning to obtain the path cost, so it is necessary to use GVD path cost.
When the robot reaches the target frontier, the map will be updated and local frontiers will be extracted.
At this time, the robot adopts the real-time local frontier strategy or local frontier strategy.
Therefore, it is not necessary to calculate all the global frontiers, and the global frontier strategy only determines which area the robot goes to explore.
Here, we cluster the global frontiers to reduce computational consumption.
Once the optimal frontier is determined, the frontier assignment will be closed until the robot reaches the target.
When $V_{current}$, $V_{local}$ and $V_{global}$ are all empty, it means that there are no exploration frontiers extracted on the map at this time, and the robot exploration ends.

\begin{figure}[]
	\subfigure[]{
		\centering
		\includegraphics[width=0.234\textwidth]{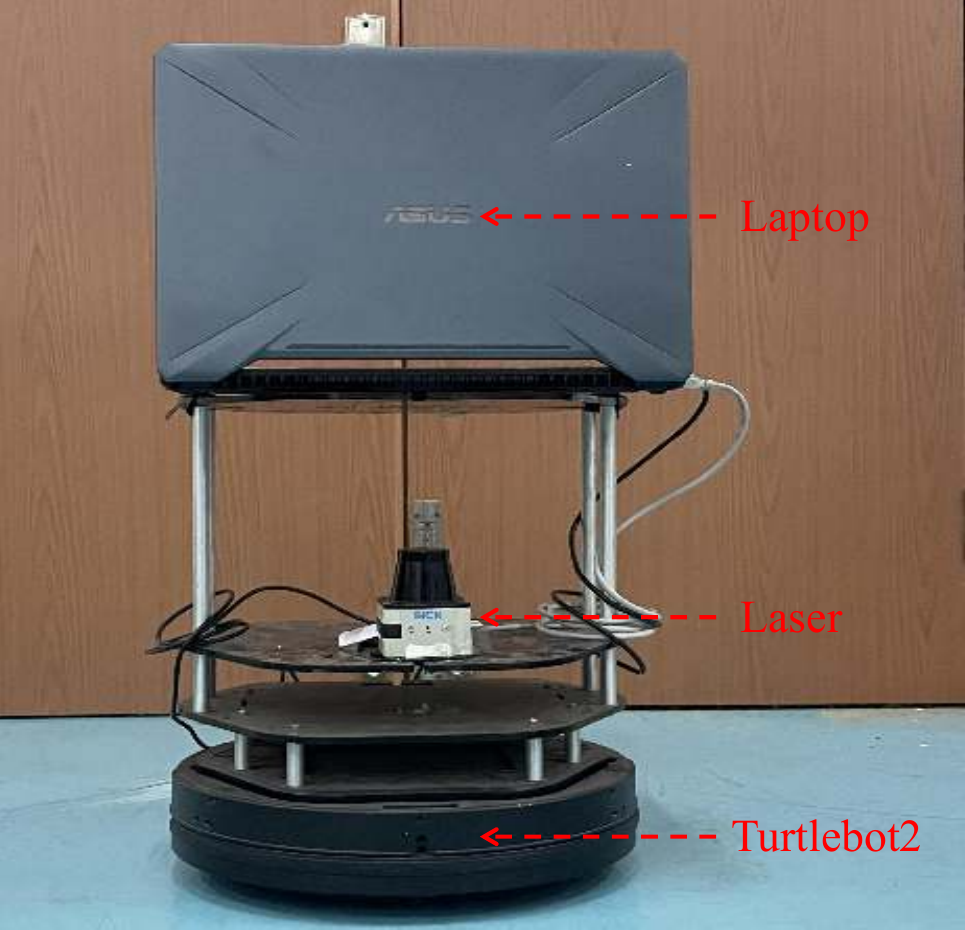}
	}\subfigure[]{
		\centering
		\includegraphics[width=0.065\textwidth]{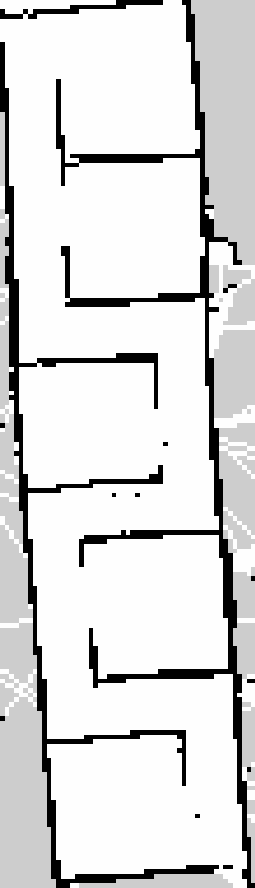}
	}\subfigure[]{
		\centering
		\includegraphics[width=0.125\textwidth]{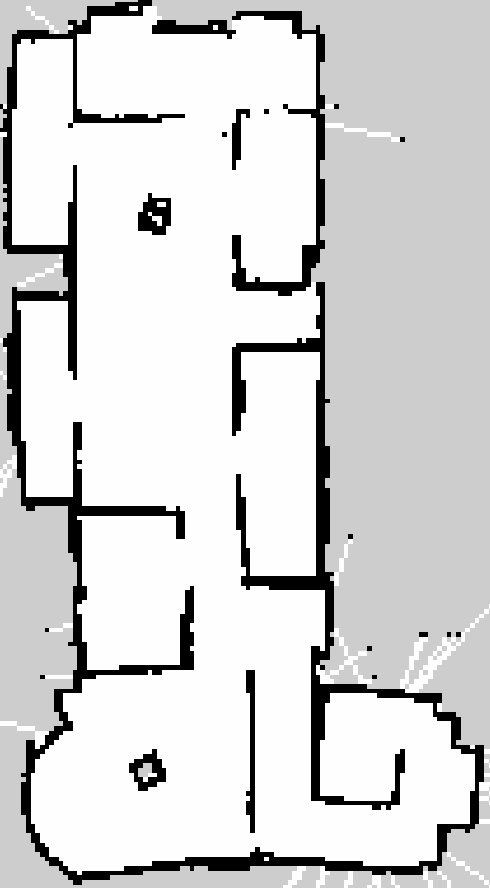}
	}
	\caption{Robot platform in the real-world experimental studies. (a) Robot platform. (b) and (c) Real-world environments.}
	\label{fig55}
\end{figure}

%机器人的图片我放入到pic文件夹里了，但我感觉单放一个不合适，而且就一个机器人，一个雷达，一台电脑，我感觉不是很有必要
\section{EXPERIMENTAL STUDIES AND RESULTS}
\label{sec:results}
\subsection{Experiment Setup}
To verify the effectiveness and superiority of our proposed GVD-Exploration method, we conduct experiments in six different types of simulated environments and two real-world environments according to the complexity of the environments.
%The map size of simulated and real-world environments are shown in Table \ref{tbl:table3}.
We carry out the experiments on the basis of Ubuntu 16.04 operating system and Robot Operating System (ROS).
The experimental scene of the simulation environments are built by Gazebo.
%The map size of the simulation environment is shown in Fig. \ref{fig8}.
For the simulation environment, the sensing distance and the sensing angle of the laser sensor are 10${m}$ and 270$^{\circ}$, respectively.
For the real environment of the robot autonomous exploration, we adopt the Turtlebot2 equipped with a sick-561 laser sensor as the experimental platform, as shown in Fig. \ref{fig55}.
The laser sensor is 10${m}$ and the sensing distance is 270$^{\circ}$.
The mobile robot Turtlebot2 has a maximum linear velocity of 0.3 ${m/s}$ and an angular velocity of 2.0 ${rad/s}$.
In both the simulated and real-world environments, the resolution of the environment map explored by robot autonomous exploration is 0.1 meters/pixel.
To ensure the fairness of the experiment, the simulations and real-world experiments of all comparison methods are operated on the same computer equipped with AMD i7-9750H CPU and 8G memory.
We repeated five trials for each group in simulations and three trials in real-world experiments.
Three methods are adopted as reference, including the multi-RRT exploration (Multi-RRT) \cite{18}, nearest exploration (Nearest) \cite{16}, and greedy exploration (Greedy) \cite{24}.
\begin{itemize}
	\item
	\textbf{Multi-RRT:} the robot evaluates frontiers based on the information gain of frontier itself and the Euclidean distance from the robot to the frontier, and selects the frontier with the highest evaluation as the goal.
	\item
	\textbf{Nearest:} the robot evaluates frontiers by the Euclidean distance from itself to the frontiers, and chooses the frontier with the smallest distance as the goal.
	\item
	\textbf{Greedy:} the robot evaluates frontiers by the information gain of the frontiers, and the frontier with the largest information gain is selected as the goal.
\end{itemize}

\begin{figure}[]
	\subfigure[sampling points = 2000]{
		\centering
		\includegraphics[width=0.23\textwidth]{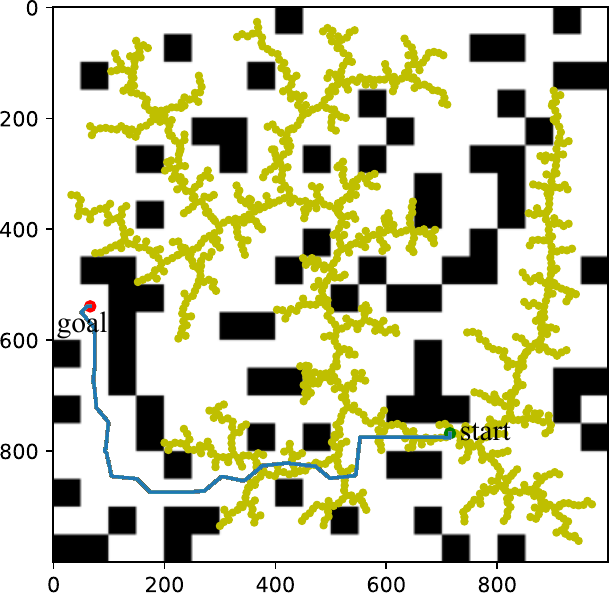}
	}\subfigure[sampling points = 5000]{
		\centering
		\includegraphics[width=0.23\textwidth]{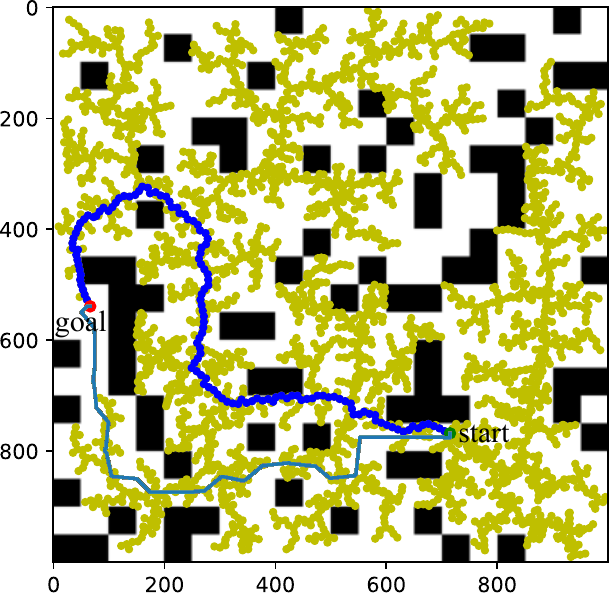}
	}\caption{Path planning based on GVD and Infromed-RRT*. The teal path is GVD planning path, and blue path is Infromed-RRT* planning path.}
	\label{fig6}
\end{figure}

\subsection{Path Planning Experiment}
The Rapidly-exploring Random Tree (RRT) \cite{4} have gained popularity for their capability of efficiently searching the state space.
Among the variants of RRT, Informed RRT* retains the same probabilistic guarantees of completeness and optimality as RRT* while improving the convergence speed and the quality of the final solution \cite{35}.
To illustrate the efficiency of the proposed GVD path planning method, we first compare our method with Infromed-RRT* \cite{35} path planning.

Fig. \ref{fig6} shows the path planned by GVD and the path planned by Informed RRT*, respectively.
In Fig. \ref{fig6} (a), the time cost of our algorithm to construct the GVD of the environment is 0.11$s$.
Then we associate the start and goal with the nearest edge of the Voronoi diagram, and use the Dijkstra \cite{31} search algorithm to plan the path, with a time cost of 0.03$s$.
Therefore, the total planning time cost of our algorithm is 0.14$s$.
When the number of sampling points is set to 2000, Informed RRT* fails to obtain the path, and the planning fails.
In Fig. \ref{fig6} (b), the time cost of GVD construction is 0.12$s$ and the planning time cost is 0.03$s$, so the total time cost is 0.15$s$.
When the number of sampling points raises to 5000, Informed RRT* needs 9.69$s$ to obtain the path.
Obviously, the path planned by our method is faster and is not limited by the number of sampling points.
In addition, within the same map, our algorithm only needs to construct the GVD once, which further reduces the time cost of the path planning during exploration.

\subsection{Performance in Simulations}
In this part, we use the GAZEBO simulation platform to build six environmental maps for robot autonomous exploration.
We compare the effectiveness of different methods in the process of robot autonomous exploration from the following three indicators: 1) the robot trajectory, 2) the map entropy, and 3) the time cost and path cost of exploring the entire environment.

\begin{figure*}[]
	\centering
	\includegraphics[width=0.78\textwidth,height=0.92\textwidth]{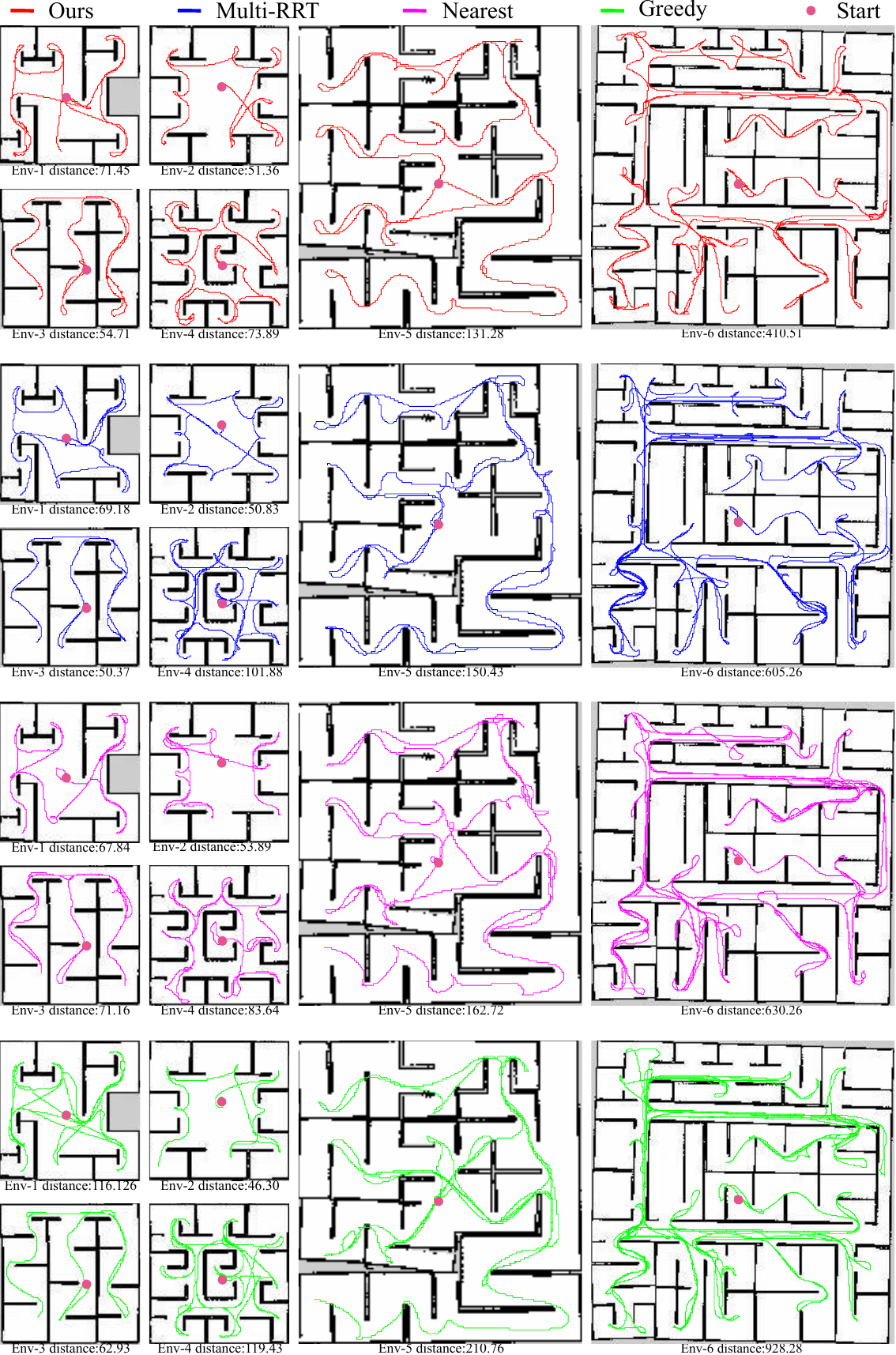}
	\caption{Robot exploration trajectories of different methods in all simulation environments.}
	\label{fig8}
\end{figure*}

\subsubsection{The trajectory during the exploration process}
Fig. \ref{fig8} shows the exploration trajectories of different methods.%and below each environment the path cost of the method in the environment is shown.
It can be seen that in simple environments (namely $Env$-$1$, $Env$-$2$ and $Env$-$3$), our method go deeply into the environments to guarantee sufficient exploration and the high quality of the environment mapping.
This strategy may not achieve the shortest path length in these simple environments, but we believe that this path loss is acceptable and can effectively guarantee the integrity of the exploration.
%Due to the sensor sometimes cannot completely cover all the unknown areas during the exploration process, the robot will turn to other areas for exploration due to the untimely extraction of frontiers, which will cause the backtracking during the robot exploration process.
Due to the sensor sometimes being unable to completely cover all the unknown areas during the exploration process, the robot may turn to other areas for exploration because of the untimely extraction of frontiers, leading backtracking during the robot exploration process.
When the complexity of the environment increases, %the robot needs to comprehensively consider different frontiers exploration sequences to speed up the construction of the environment map.
the exploration trajectory of the robot based on our method is concise and clear, and the trajectory coincidence in public area of during exploration process is lower, as shown in $Env$-$4$, $Env$-$5$ and $Env$-$6$.
Our method guarantees that the robot can fully explore along one exploration branch after another.
%After fully exploring the current branch, the robot select other branches in the current area, which ensures that the area can be fully explored.
For multiple areas, %the robot determines the order of exploration of each area by combining the information gain of the area and the path cost of all areas explored.
our method seeks the optimal exploration path as a whole other than maximizing the instant rewards.
The other methods we compared exhibit greater trajectory redundancy in certain common areas.
These methods tend to prioritize selecting the optimal exploration frontier at the current moment, disregarding the continuity between frontiers.
Because of their short-sighted nature, these methods limit the efficiency of robot autonomous exploration, which accounts for the backtracking phenomenon observed.
%The other methods we compared rely on greedy strategies, where efficiency is limited due to the methods being myopic, which leads to severe backtracking during robot exploration.
In addition, the robot may mistakenly choose a target frontier just on the other side of a wall while evaluating optimal frontiers, which is a result of inaccurate assessment of path costs.
Euclidean distance does not consider the connectivity of the environment, which often makes the target evaluation inaccurate.
Therefore, the exploration trajectories of these methods are multi-segmented and overlapping.

\begin{figure*}[ht!]
	\centering
	\includegraphics[width=6.6in]{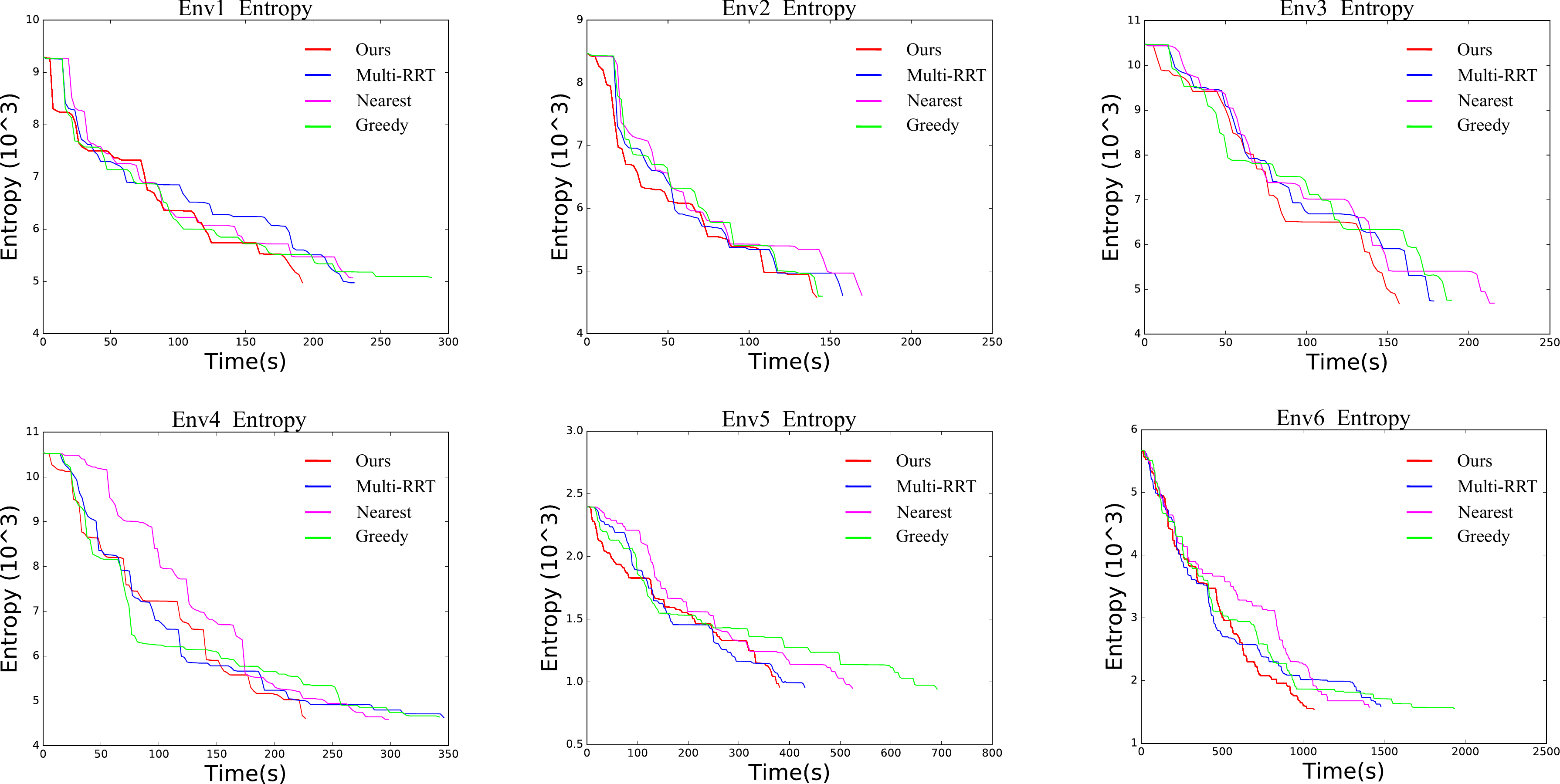}
	\caption{Map entropy of different methods during the robot exploration in all simulation environments.}
	\label{fig9}
\end{figure*}

\subsubsection{The curve of map entropy over time}
Fig. \ref{fig9} shows the curve of map entropy versus exploration time for different methods.
Among all methods, our method has the fastest decrease in map entropy, which means that our method is better than the other three methods in terms of exploration speed.
%In Env 2, our method entropy is about the same as Greedy, but our method entropy decreases faster.
%Except Env 2, our method entropy decay has a more obvious advantage, which also shows the effectiveness of our method.
It is noteworthy that in the early stages of exploration, our map entropy decline is usually not the fastest since the immediate information gain is not the main consideration factor in our cost function.
%Compared with our methods, those other methods show that the entropy stops with time during the exploration, which indicates that the robot exploration process may backtrack or the robot is trapped in place without frontiers.
%The information entropy of our method decays continuously, which indicates frontiers extraction timely and less exploration backtracking.

\begin{center}
	\begin{table*}[ht!]
		\centering
		\caption{EXPERIMENTAL DATA ON ROBOT EXPLORATION IN SIMULATED ENVIRONMENTS}
		\setlength{\tabcolsep}{7mm}{
			\begin{tabular}{cccccc}
				\toprule
				\multirow{2}{*}{\textbf{Map Index}} & \multirow{2}{*}{\textbf{Map Size}}& \multicolumn{4}{c}{\textbf{Methods}} \\
				%\textbf{Map Index} & \multicolumn{2}{c}{\textbf{Map Size}} & \multicolumn{4}{c}{\textbf{Methods}} \\
				\cline{3-6} 
				\textbf{} & \textbf{} & \textbf{Multi-RRT} & \textbf{Nearest} & \textbf{Greedy} & \textbf{Ours} \\
				\midrule
				\textbf{Time Cost(s)} \\
				$Env$-$1$ & 10m x 10m & 236.48(17.49\%)   & 224.01(12.90\%)  & 304.35(35.88\%)  & \textbf{195.11}  \\
				$Env$-$2$ & 10m x 10m & 156.56(9.22\%)    & 167.32(15.05\%)  & 151.48(6.17\%)   & \textbf{142.13}  \\
				$Env$-$3$ & 10m x 10m & 199.33(22.59\%)   & 217.27(28.98\%)  & 253.69(39.18\%)  & \textbf{154.30}  \\
				$Env$-$4$ & 10m x 10m & 323.19(30.64\%)   & 296.19(24.30\%)  & 385.77(41.89\%)  & \textbf{224.17}  \\
				$Env$-$5$ & 16m x 16m & 457.13(15.06\%)   & 496.29(21.76\%)  & 635.57(38.91\%)  & \textbf{388.29}  \\
				$Env$-$6$ & 30m x 30m & 1474.73(12.92\%)  & 1483.37(13.42\%) & 1983.21(35.24\%) & \textbf{1284.26} \\
				
				\midrule
				\textbf{Path Cost(m)} \\
				$Env$-$1$ & 10m x 10m & 72.56(3.18\%)    & 71.26(1.42\%)   & 107.82(34.85\%) & \textbf{70.25}  \\
				$Env$-$2$ & 10m x 10m & 50.14(-1.97\%)   & 49.75(-2.77\%)  & 52.27(2.18\%)   & \textbf{51.13}  \\
				$Env$-$3$ & 10m x 10m & 54.64(-3.04\%)   & 67.05(16.02\%)  & 81.36(30.79\%)  & \textbf{56.31}  \\
				$Env$-$4$ & 10m x 10m & 96.62(23.16\%)   & 88.09(15.72\%)  & 123.98(40.12\%) & \textbf{74.24}  \\
				$Env$-$5$ & 16m x 16m & 152.57(6.38\%)   & 161.38(11.49\%) & 227.56(37.23\%) & \textbf{142.83} \\
				$Env$-$6$ & 30m x 30m & 536.65(15.79\%)  & 513.76(12.04\%) & 774.50(41.65\%) & \textbf{451.91} \\
				\bottomrule
		\end{tabular}}
		\label{tbl:table1}
		\centering
	\end{table*}
\end{center}

\begin{figure}[ht!]
	\centering
	\subfigure[Time cost comparison of different method.]
	{
		\centering
		\includegraphics[width=3.3in]{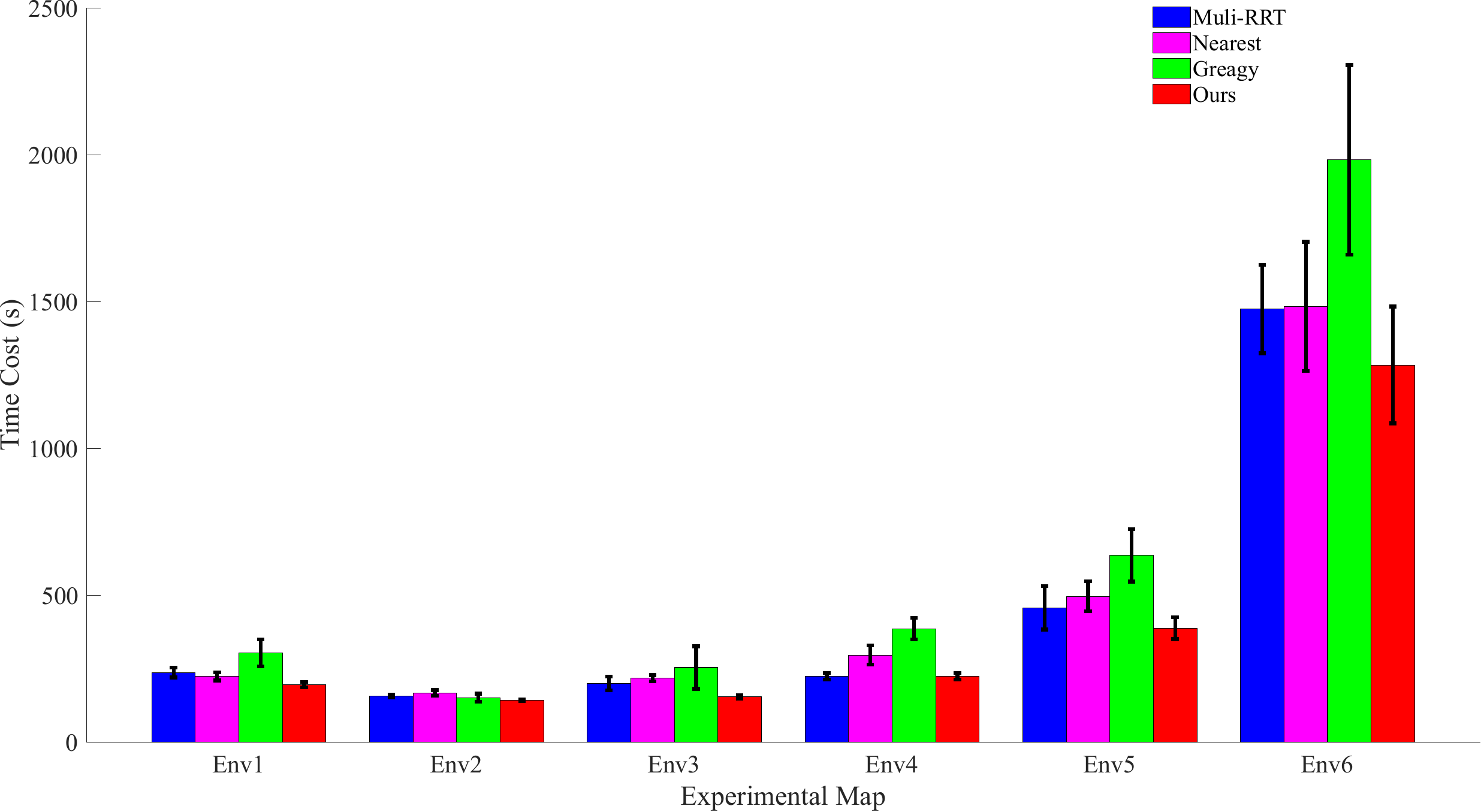}
	}
	\subfigure[Path cost comparison of different method.]
	{
		\centering
		\includegraphics[width=3.3in]{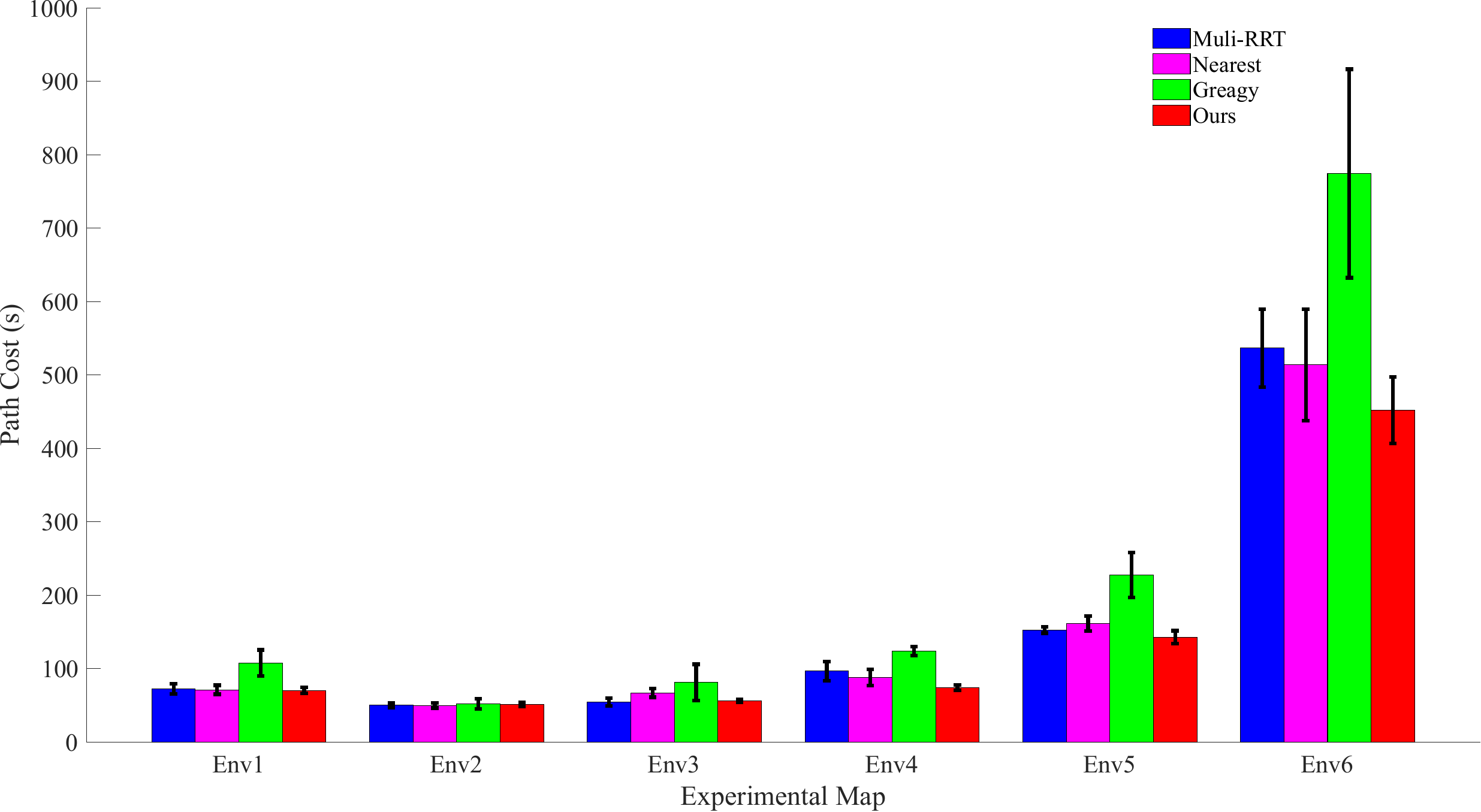}
	}
	\caption{The time and path cost of robot exploration in simulation environments.}
	\label{fig10}
\end{figure}

\subsubsection{The time cost and path cost of the exploration}
The exploration time cost and path cost of the four methods are shown in Fig. \ref{fig10}.
It can be found that our method has the lowest time cost in all environments, which reveals the efficiency of our method.
For the path cost, in simple environments such as $Env$-$1$, $Env$-$2$ and $Env$-$3$, our path cost is also close to optimal.
Although our method sacrifices some path cost to improve the mapping quality, it is still possible accepted.
For some complex environments, such as $Env$-$4$, $Env$-$5$ and $Env$-$6$, it is obvious that our method has the lowest path cost, which further shows the effectiveness of our frontier selection strategy.
The Multi-RRT method and the Nearest method are comparable in path cost, but the Multi-RRT method is superior to the Nearest method in terms of time cost, which indicates that compared with a single path cost, comprehensive consideration of the information gain and path cost of the frontier helps in  accelerating the robot exploration.
The Greedy method has the highest cost in terms of both time and path cost and it is prone to backtracking during the exploration process. \\

\begin{figure*}[ht!]
	\centering
	\includegraphics[width=6.6in]{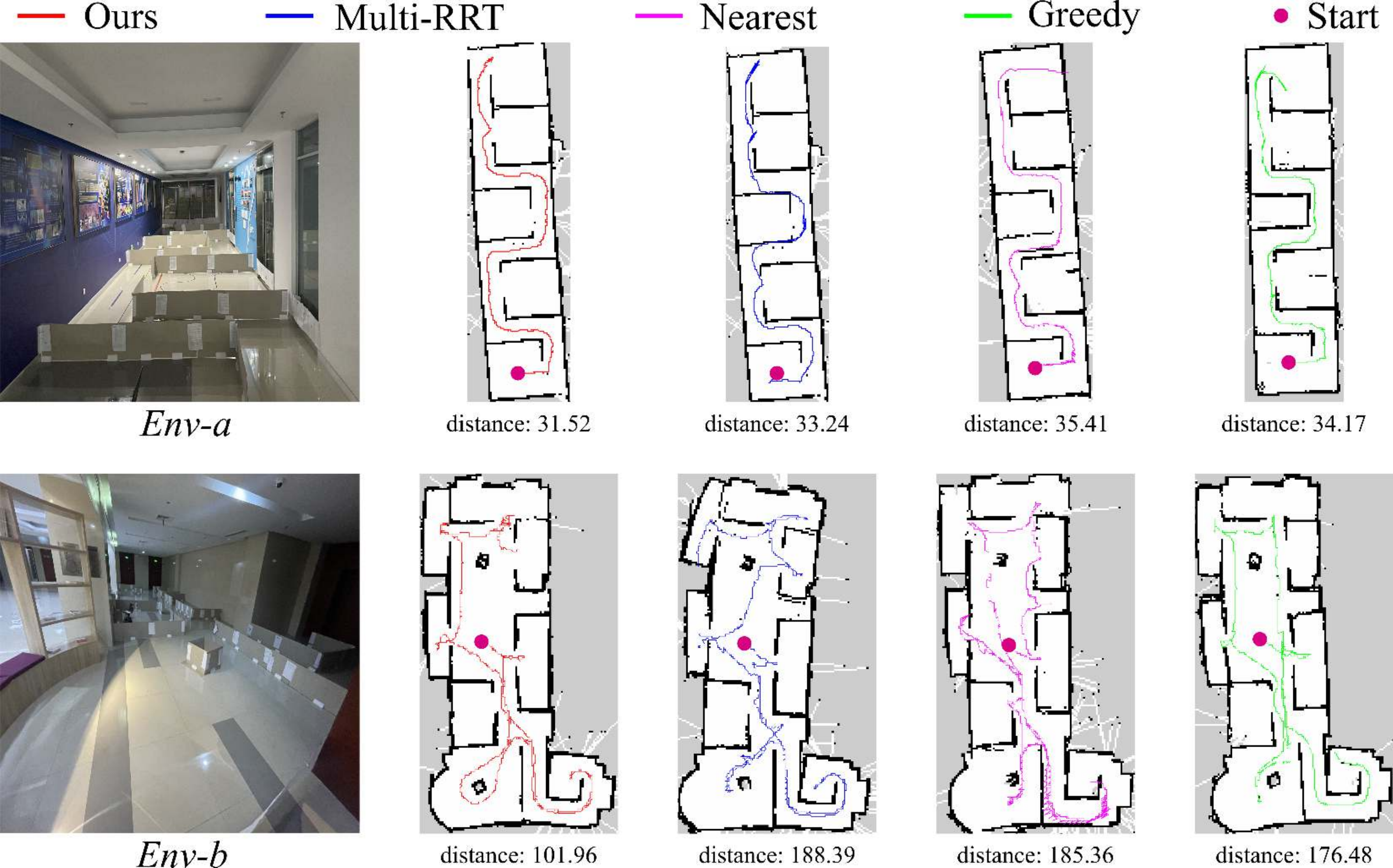}
	\caption{Robot exploration trajectories of different methods in real-world environments.}
	\label{fig11}
\end{figure*}

%We also recorded the average path cost and time cost of the four methods in all environments and performed percentage quantitative analysis, as shown in Table \ref{tbl:table1}.

We also recorded the average path cost and time cost of the four methods in all environments and conducted a quantitative analysis in terms of percentages., as shown in Table \ref{tbl:table1}.
The numbers presented in the table represent the mean values, with parentheses indicating the corresponding percentage differences from our method.
Compared with other three methods, our method achieves over 10$\%$ improvement in time cost except for $Env$-$2$ (over 5 $\%$ ).
%For Env 2, our method also has the advantage of time cost.
%This means that our method can allow robots to perform fast and efficient exploration in any environment.
For path cost, our method is slightly inferior to the Muli-RRT method and the Nearest method in $Env$-$2$ and $Env$-$3$.
For other environments, our method obtains obvious advantage in path cost, which further validates the effectiveness of our method.
%Therefore, from the analysis of time cost and path cost, we think our method is completely feasible.

\subsection{Performance in Real-world Experiments}
In this work, we adopt two real environments to verify the effectiveness of our method, as shown in the Fig. \ref{fig55}.
$Env$-$a$ is a one-way exploration environment to verify the influence of frontiers extraction speed during robot autonomous exploration.
$Env$-$b$ is a multi-choice structured exploration environment to verify the rationality of our frontier assignment strategy.
The exploration trajectories of the robot in the corresponding scene using different methods are also given, as shown in the Fig. \ref{fig11}.
%We mainly evaluate each method in terms of robot exploration trajectory and time and path cost.
%The exploration time cost and path cost of the four methods in real environments are shown in Fig. \ref{fig12}.
We recorded the average path cost and time cost of the four methods in real environments and performed percentage quantitative analysis, as shown in Table \ref{tbl:table2}.

%\begin{figure}[]
%\centering
%\subfigure[Time cost comparison of different method.]
%{
	%        \centering
	%        \includegraphics[width=3.3in]{pic/P32realtime.eps}
	%}
%\subfigure[Path cost comparison of different method.]
%{
	%        \centering
	%        \includegraphics[width=3.3in]{pic/P33realdis.eps}
	%}
%\caption{The time and path cost of robot exploration in real environments.}
%\label{fig12}
%\end{figure}

Since the $Env$-$a$ is a one-way environment, the path length is not much different, and we mainly compare the methods in terms of time cost.
In the real-world experiments, the advantages of the proposed method is more obvious, which achieves over 30$\%$ improvements in time cost.
%Our method has a big advantage in time, at least 30$\%$, which shows that our method can quickly extract frontiers in a complex and narrow environment to guide the robot to conduct autonomous exploration.
It can be observed that $Env$-$a$ also contains several narrow corridors and corners, which makes it difficult for the growth of the RRT Tree and the RRT-based frontiers extraction.  
Therefore, the robot may occasionally be trapped in place, resulting in prolonged exploration time.
Or even worse, they cannot finish the exploration of the entire environment within reasonable time cost, resulting in a failure.
%From the exploration time, we think our method is performing better.

\begin{center}
	\begin{table*}[]
		\centering
		\caption{EXPERIMENTAL DATA ON ROBOT EXPLORATION IN REAL ENVIRONMENTS}
		\setlength{\tabcolsep}{7mm}{
			\begin{tabular}{cccccc}
				\toprule
				\multirow{2}{*}{\textbf{Map Index}} & \multirow{2}{*}{\textbf{Map Size}}& \multicolumn{4}{c}{\textbf{Methods}} \\
				%\textbf{Map Index} & \multicolumn{2}{c}{\textbf{Map Size}} & \multicolumn{4}{c}{\textbf{Methods}} \\
				\cline{3-6} 
				\textbf{} & \textbf{} & \textbf{Multi-RRT} & \textbf{Nearest} & \textbf{Greedy} & \textbf{Ours} \\
				\midrule
				\textbf{Time Cost(s)} \\
				$Env$-$a$ & 4.4m x 21m & 459(53.96\%)   &506(58.24\%)  & 319.33(33.82\%)  & \textbf{211.33}  \\
				$Env$-$b$ & 18.3m x 13m & 581.33(13.47\%)    & 714(29.55\%)  & 637.67(21.12\%)   & \textbf{503}  \\
				\midrule
				\textbf{Path Cost(m)} \\
				$Env$-$a$ & 4.4m x 21m & 33.24(5.17\%)    & 35.41(10.99\%)   & 34.17(7.76\%) & \textbf{31.52}  \\
				$Env$-$b$ & 18.3m x 13m & 188.39(45.88\%)   & 185.36(44.99\%)  & 176.48(42.23\%)   & \textbf{101.96}  \\
				\bottomrule
		\end{tabular}}
		\label{tbl:table2}
		\centering
	\end{table*}
\end{center}

There are several distributed rooms in $Env$-$b$, and the difficulty lies in target frontier decision-making to improve the exploration efficiency.
%From the exploration trajectory, our method trajectory is clearer, indicating that the robot exploration is more efficient.
Analyzing the exploration trajectory, it becomes evident that the trajectory of our method is more distinct, indicating enhanced efficiency during the robot exploration.
Our method has achieved over 40$\%$ improvement in path length and over 10$\%$ improvement in exploration time cost.
In a complex environment, the greedy and short-sighted decision-making will cause the robot to continuously backtrack during the exploration process, resulting in an increase in path length and greater resource consumption.
Our method can not only ensure that the robot continues to explore along the exploration direction, but also comprehensively consider the exploration sequence between the frontiers.
Therefore, starting from the path length, our frontiers assignment strategy is more competitive.

\section{CONCLUSIONS}
\label{sec:conclusions}

In this paper, a Generalized Voronoi Diagram (GVD) based framework with multiple choice strategies has been proposed for robot autonomous exploration.
The method includes a new novel mapping model to accelerate GVD construction, heuristic frontiers generation and multi-strategy allocation of frontiers.
Different from previous work \cite{6}, the new mapping model has better real-time performance, reducing computational complexity, and is more suitable for autonomous exploration by mobile robots.
For heuristic frontiers extraction, in addition to the global frontiers, we increase the extraction of local frontiers.
When assigning frontiers, for local frontiers in the local area of the robot, we use path cost and information gain for joint evaluation, and for global frontiers, we solve the exploration order of frontiers by solving the TSP problem.
Through a series of simulations and real experiments, the results show that compared with other RRT-based methods, our method has more advantages in time cost and path cost, and our method has higher exploration efficiency.
In the future, we want to extend this strategy to multi-robot autonomous exploration, which requires coordinating the order of exploration between robots and we think it is also challenging.

\bibliographystyle{IEEEtran}
\bibliography{reference.bib}

\end{document}